\documentclass[letterpaper, 10 pt, conference]{ieeeconf}  
\IEEEoverridecommandlockouts
\usepackage{cite}
\usepackage{amsmath,amssymb,amsfonts}
\usepackage{algorithmic}
\usepackage{graphicx}
\usepackage{textcomp}
\usepackage{xcolor,color,soul}
\usepackage{booktabs}
\usepackage{xurl}
\usepackage{float}
\usepackage{stfloats}
\usepackage[hidelinks]{hyperref}
\def\BibTeX{{\rm B\kern-.05em{\sc i\kern-.025em b}\kern-.08em
    T\kern-.1667em\lower.7ex\hbox{E}\kern-.125emX}}
\makeatletter
\def\bstctlcite{\@ifnextchar[{\@bstctlcite}{\@bstctlcite[@auxout]}}
\def\@bstctlcite[#1]#2{\@bsphack\@for\@citeb:=#2\do{\edef\@citeb{\expandafter\@firstofone\@citeb}\if@filesw\immediate\write\csname #1\endcsname{\string\citation{\@citeb}}\fi}\@esphack}
\makeatother
\begin{document}
\bstctlcite{IEEEexample:BSTcontrol}
\raggedbottom

\title{\LARGE \bf Ego2Contact: Predicting Touch from Egocentric Human Video for Contact-Rich Robot Manipulation
}

\author{%
\parbox{\textwidth}{\centering
Qianqian~Yang$^{1}$,
Dayou~Li$^{2}$,
Bangya~Liu$^{3}$,
Zihao~Zhu$^{2}$,
Nuo~Chen$^{2}$,
Siyuan~Xu$^{2}$,
Hezhen~Hu$^{4}$,
Chen Yu$^{4}$,
Rebecca~Friesen$^{2}$,
Minghui~Zheng$^{2}$,
Cynthia~Hipwell$^{2}$,
Xiaoning~Qian$^{2}$,
Kenji~Shimada$^{1}$,
and Zhiwen~Fan$^{2}$
}%
\thanks{Corresponding author: Zhiwen Fan
(e-mail: zhiwenfan@tamu.edu).}%
\thanks{$^{1}$Carnegie Mellon University, Pittsburgh, PA, USA.}%
\thanks{$^{2}$Texas A\&M University, College Station, TX, USA.}%
\thanks{$^{3}$UW-Madison, Madison, WI, USA.}%
\thanks{$^{4}$UT Austin, Austin, TX, USA.}%
}

\maketitle
\thispagestyle{empty}
\pagestyle{empty}

\begin{abstract}
Inferring touch from diverse human interaction videos offers a way to learn visual--tactile priors for robot manipulation. In contrast, collecting robot tactile data requires specialized hardware and platform-specific demonstrations, which limits its scale. Existing predictors learn the vision-to-touch mapping from paired human data, but gaps in hand appearance and tactile layout prevent them from directly supplying touch to robot policies. To address this limitation, we present Ego2Contact, which learns a visual--tactile prior from egocentric human interactions and transfers it to robot hands as tactile observations for policy learning. Its predictor fuses temporal scene context with fine-grained hand-local visual features to estimate bimanual tactile fields from RGB video. To bridge both gaps, we adopt a canonical grid as a unified tactile representation and map varying layouts of tactile gloves and robot hands onto it through anatomical region correspondences, turning robot measurements into fine-tuning targets and predictions into policy inputs. Experiments show that Ego2Contact improves tactile prediction accuracy over previous work on human data and simulated robot data after adaptation. The transferred prior further benefits policy learning. Compared with training without tactile input, predicted touch raises the average success rate from 44.4\% to 48.5\% for reinforcement learning on a simulated Shadow hand with link-level force sensing, and from 27.5\% to 45.0\% for policy post-training on real robot hands wearing dense taxel gloves. Egocentric human interactions could thus serve as a source of visual--tactile priors for robot manipulation when tactile measurements are limited.

\end{abstract}


\section{Introduction}
Reliable robot manipulation requires understanding both the scene captured in the video and the physical interaction within it, including where contact occurs and how forces evolve~\cite{johansson2009tactile}. Tactile sensing provides direct measurements of this contact to supplement video~\cite{calandra2018feeling}, yet collecting large-scale synchronized visual--tactile pairs on robots remains challenging. Tactile skins and optical sensors require specialized hardware that is difficult to integrate~\cite{dahiya2010tactile,gelsight}, so demonstrations of contact-rich dexterous behaviors are costly to collect. These recordings are also tied to the geometry and sensor layout of a specific hand, limiting the transfer across various hardware platforms.

\begin{figure*}[t]
    \centering
    \includegraphics[width=0.98\textwidth]{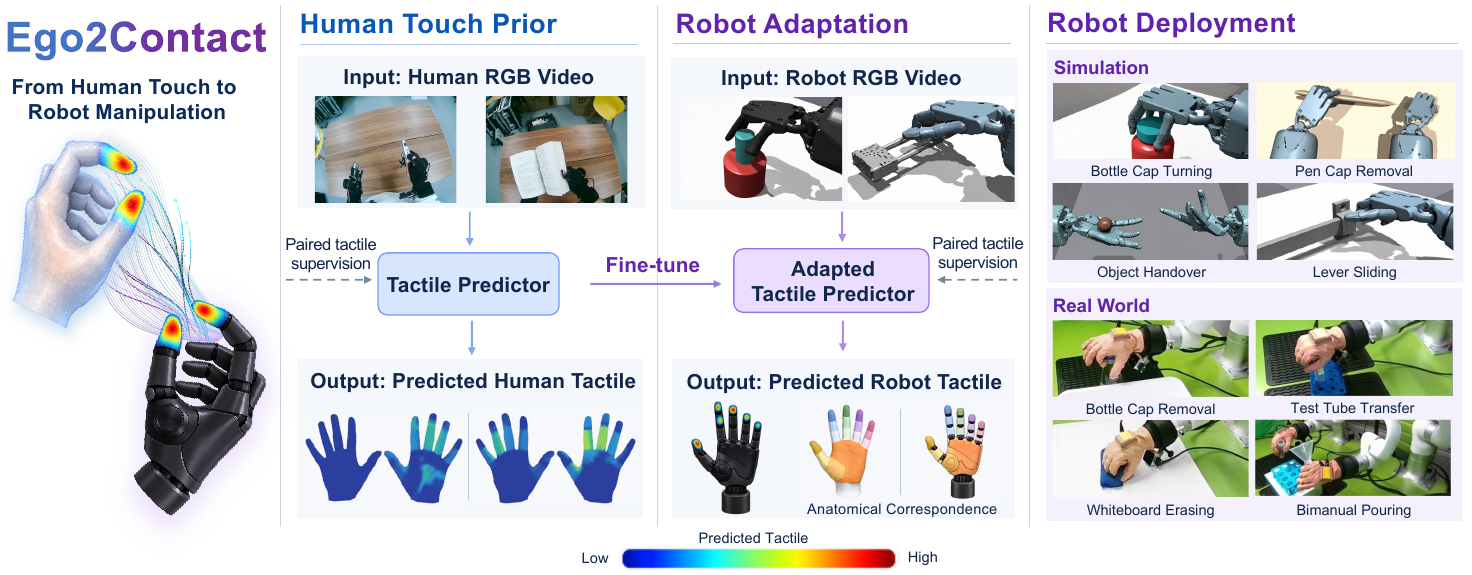}
    \vspace{-4mm}
    \caption{\textbf{From human touch to robot manipulation.} We learn a visual--tactile predictor from paired human egocentric videos and tactile measurements, fine-tune it to the dexterous robot hand domain through anatomical region correspondences, and use the adapted predictor to supply touch for downstream policy learning across four simulated and four real-world tasks.}
    \label{fig:teaser}
\end{figure*}

By contrast, human egocentric video is considerably easier to collect at scale and covers diverse everyday interactions~\cite{epickitchens,ego4d,egodex}, but rarely includes tactile signals. Recent datasets pair RGB video with hand pose~\cite{egodex} or dense full-hand pressure from sensing gloves~\cite{opentouch,touchanything,egotactile}. Such glove data still requires an instrumented hand for every interaction, limiting how far it can scale. If touch can instead be inferred from video, a predictor trained on glove data could supply tactile signals to videos lacking them, including robot demonstrations without tactile sensors.

Existing visual--tactile predictors learn touch from paired human data but are developed and evaluated within the human domain~\cite{pressurevision,touchanything,egotactile}, leaving open whether the inferred touch can drive robot policies. Robot-centric methods such as TacImag~\cite{tacimag} and TouchWorld~\cite{touchworld} avoid the human-to-robot transfer by predicting touch within the robot domain, but they rely on paired robot data or measured touch inputs, inheriting the same collection cost. Methods that transfer human tactile priors to robots connect the two domains~\cite{vtdexmanip,htac}, but they entangle tactile estimation with a specific policy architecture or action space, so the learned prior is hard to reuse across robots and policies. What remains missing is a tactile predictor that learns from human video and delivers its outputs as explicit tactile observations to policies on robot hands with different sensor layouts.

Building such a predictor raises two fundamental challenges. The first is visual ambiguity: contact regions occupy only a small fraction of each frame, and the same hand pose can exert vastly different force patterns across task phases. Reliable estimation therefore needs fine-grained hand appearance together with temporal scene context. The second is the embodiment gap in visual appearance and tactile layout: robot hands look different from gloved human hands, and their tactile signals range from sparse link-level forces in simulation to dense taxel arrays on real hardware. Sharing one predictor across robot hands requires a unified tactile representation that maps diverse layouts, allowing the human-pretrained model to adapt to each hand.

To address these challenges, we introduce \textbf{Ego2Contact}, a framework that learns a visual--tactile prior from human egocentric video and transfers it to robot hands, providing tactile observations for policy learning (Figure~\ref{fig:teaser}). To reduce visual ambiguity, its predictor fuses temporally aggregated scene context with fine-grained hand-local visual features through cross-attention. A conditional flow-matching generator then turns the fused features into bimanual tactile fields on a canonical hand grid partitioned into anatomical regions. To bridge the embodiment gap, we link each robot's sensing zones to these anatomical regions, mapping robot tactile recordings onto the grid as targets for fine-tuning on robot video and decoding grid predictions into policy inputs. This shared representation lets the human-pretrained predictor adapt across diverse kinematics and sensors without changing its output space. Its outputs are explicit tactile observations, so it also stays decoupled from the downstream policy and serves both reinforcement learning and vision-tactile-language-action (VTLA) post-training.

\begingroup
\setlength{\parskip}{2pt}
We evaluate Ego2Contact from tactile prediction to robot policy learning. In tactile prediction, it outperforms regression and pose-conditioned baselines on most metrics for seen and unseen objects in human data, and after adaptation it localizes contact more accurately than the regression baseline on simulated robot data. We then transfer the predictor to two robot hands with different tactile layouts. On a simulated Shadow hand with 17 link-level force sensors, it supplies contact features online for reinforcement learning. On real robot hands wearing 880-taxel gloves, it annotates demonstration videos with dense tactile images for VTLA post-training. In both settings, policies using predicted touch outperform those trained without tactile input and close part of the gap to those trained with ground-truth touch. Additional prediction results and robot rollouts are available in the supplementary material.

In summary, this work has the following contributions.
\begin{itemize}\setlength{\itemsep}{2pt}\setlength{\parsep}{0pt}
    \item We propose Ego2Contact, a framework that transfers a visual--tactile prior learned from human egocentric video to robot hands, where it supplies explicit tactile observations for policy learning.
    \item We introduce a conditional flow-matching tactile predictor that fuses temporal scene context with hand-local visual features, paired with anatomical correspondences mapping diverse sensor layouts to a canonical grid for robot adaptation and policy input.
    \item We evaluate Ego2Contact from human tactile prediction to reinforcement learning in simulation and VTLA post-training on real robots, showing that predicted touch improves manipulation success in both settings.
\end{itemize}
\par\endgroup

\begin{figure*}[t!]
    \centering
    \includegraphics[width=0.96\textwidth]{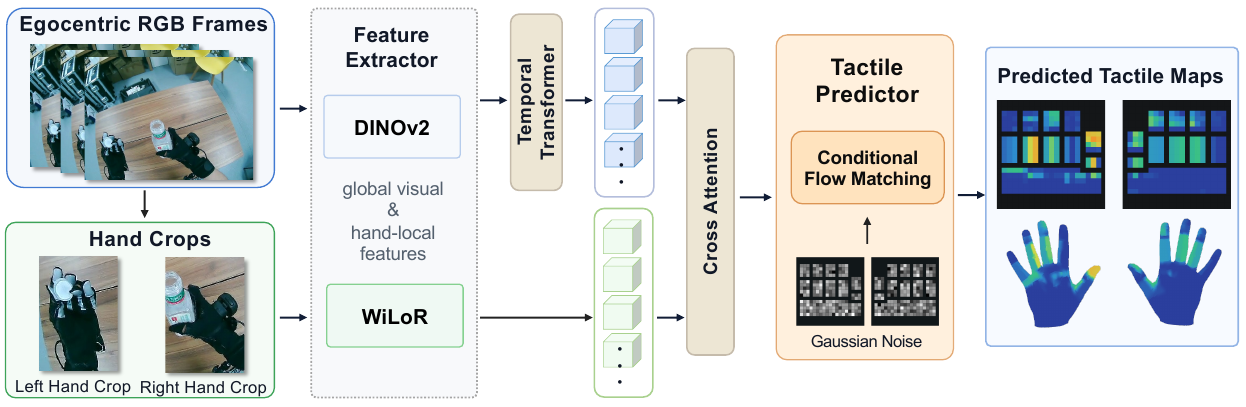}
    \vspace{-3mm}
    \caption{\textbf{Visual--tactile prediction model.} Frozen DINOv2 and WiLoR backbones extract scene and hand-local features. Temporal aggregation and cross-attention condition a flow-matching generator that produces bimanual $21\times21$ tactile fields from Gaussian noise at inference. During training, the generator receives ground-truth fields interpolated with noise and learns the corresponding velocity field.}
    \label{fig:human_touch_prior}
\end{figure*}

\section{Related Work}
\subsection{Vision-Based Contact and Pressure Estimation}
Predicting physical interaction from images has progressed from localizing contact to estimating continuous pressure. Early datasets capture where contact occurs on object or hand surfaces~\cite{brahmbhatt2019contactdb,taheri2020grab}, and dense predictors estimate binary hand-surface contact~\cite{haco}. A parallel line of work regresses hand pressure from RGB images, progressing from planar touchpads~\cite{pressurevision,pressurevisionpp} to egocentric observations with pose and mesh-space pressure~\cite{egopressure,egopressdiff}. Recent work scales this to full-hand tactile fields by pairing egocentric video with wearable-glove measurements~\cite{touchanything,egotactile,egotac,opentouch,htbench}, and estimates contact and 3D force jointly from monocular input~\cite{egophi,hope,rlhnd}. Among these, TouchAnything introduces the EgoTouch dataset and a bimanual predictor with the canonical layout we adopt~\cite{touchanything}. These methods show that dense touch is visually predictable, but they operate and are evaluated within the human domain.

\subsection{Tactile Sensing and Imagination for Robot Manipulation}
Tactile feedback improves contact-rich manipulation when vision is ambiguous. Policies combine visual and tactile observations to select regrasping actions~\cite{calandra2018feeling} and to react to force feedback during execution~\cite{reactivediffusionpolicy}. Multimodal visual--tactile representations support contact-rich tasks~\cite{lee2019makingsense}, and tactile sensing has enabled dexterous skills such as in-hand object rotation from vision and touch~\cite{guzey2023dexterity,qi2023rotateit}. Since tactile hardware is not always available at deployment, existing work predicts tactile signals within the robot's own domain. TacImag learns tactile imagination from paired robot demonstrations for sensor-free deployment~\cite{tacimag}. ViTacWorld generates action-conditioned visual--tactile rollouts for policy training~\cite{vitacworld}, while TouchWorld integrates predicted tactile subgoals with tactile-conditioned control~\cite{touchworld}. Our anatomical interface complements these efforts by connecting human tactile predictions to different robot sensing layouts.

\subsection{Transferring Human Priors to Robot Tactile Learning}
Beyond extracting visual representations~\cite{nair2022r3m} or motion trajectories~\cite{qin2022dexmv,shaw2022videodex,wang2024dexcap} from large-scale egocentric datasets~\cite{epickitchens,ego4d,egodex}, recent efforts transfer human tactile priors to robots. VTDexManip~\cite{vtdexmanip} pretrains visual--tactile representations from human demonstrations, whereas H-Tac~\cite{htac} learns a joint tactile-action latent space transferred via post-training. However, both paradigms entangle tactile estimation with specific policy architectures or action spaces. Ego2Contact instead decouples perception from control, learning a dense pressure predictor from human video and bridging robot sensor layouts via anatomical correspondences to supply observations for simulated and real-world policy training.

\section{Methodology}
\label{sec:method}
Ego2Contact learns a visual--tactile prior from human video (Section~\ref{subsec:objectives}) and transfers it to robot hands and their policies (Section~\ref{subsec:transfer}), as shown in Fig.~\ref{fig:teaser}.

\subsection{Problem Formulation}
Let $V_t=\{I_{t-(T-1)\Delta},\ldots,I_t\}$ be an RGB clip of $T{=}8$ frames with sampling interval $\Delta{=}2$, and let $P_t\in[0,1]^{2\times21\times21}$ be the bimanual tactile field at frame $I_t$. Each hand's tactile field uses the $21\times21$ layout of~\cite{touchanything}, with 217 valid cells of normalized pressure. We first train a predictor $f_{\theta_H}$ on paired egocentric human video and tactile measurements to estimate $P_t$ from $V_t$, with human targets normalized per sequence. For each robot domain $R$, we then independently initialize $\theta_R$ from $\theta_H$ and fine-tune the predictor on robot tactile targets mapped to the same grid through anatomical region correspondences.

\subsection{Learning a Visual--Tactile Prior from Human Video}
\label{subsec:objectives}
\textbf{Visual conditioning.}
To reduce visual ambiguity, the predictor combines temporal scene context with fine-grained hand-local appearance (Fig.~\ref{fig:human_touch_prior}). To encode scene context, a frozen DINOv2 ViT-B/14 encoder~\cite{dinov2} extracts 256 patch tokens of dimension 768 from each RGB frame, and a three-layer temporal transformer aggregates them across the clip. To encode hand appearance, the frozen ViT-H backbone of the hand reconstruction model WiLoR~\cite{wilor} extracts spatial features from each hand crop, with left-hand crops mirrored to match its right-hand convention. Spatial pooling keeps 16 tokens per hand, and a learned projection maps the 1280-dimensional WiLoR features to the 768-dimensional scene-token space. These hand tokens query the temporally aggregated scene features through two cross-attention layers, producing the condition $c_t$. To keep the predictor from relying on the black glove of the human recordings, we recolor the glove with bright colors during training. Online augmentation also shifts the HSV hue with Gaussian noise separately for the hand region and the background and rotates the hand crop in plane by up to $20^\circ$. We drop the WiLoR conditioning with probability 0.2 and keep both visual backbones frozen, so tactile supervision trains only the remaining modules.

\textbf{Conditional tactile generation.}
To generate tactile fields conditioned on $c_t$, we follow DiT~\cite{dit} with a six-block transformer. Each noisy $21\times21$ hand grid is split into $3\times3$ patches and embedded as 49 tokens. Each block applies self-attention across grid locations and cross-attention to $c_t$, while zero-initialized adaptive layer normalization modulates features using the flow-time embedding and mean condition token. A shared output head then predicts a $21\times21$ velocity grid for each hand.

\textbf{Training and sampling.}
We train the generator with rectified flow~\cite{rectifiedflow}. Given a ground-truth field $P$ and Gaussian noise $\epsilon\sim\mathcal{N}(0,\mathbf{I})$, we draw $s\sim\mathcal{U}(0,1)$ and define the interpolated field $P_s$ at flow time $s$ and its velocity target $u$:
\begin{equation}
    P_s=(1-s)P+s\epsilon,\qquad u=\epsilon-P.
    \label{eq:flow_path}
\end{equation}
The generator predicts the velocity $v_\theta(P_s,s,c)$ given the visual condition $c$, and we reconstruct the clean field as $\widetilde{P}=P_s-sv_\theta(P_s,s,c)$. The loss combines MSE and $\ell_1$ terms in velocity space with total variation (TV) regularization on $\widetilde{P}$. With $e_i=v_{\theta,i}-u_i$ and the contact weight $w_i=1+(\kappa-1)\mathbf{1}[P_i>\tau_c]$ at grid cell $i$, the training objective is:
\begin{equation}
    \begin{split}
        \mathcal{L}={}&\mathbb{E}\left[
        \frac{\sum_i w_i(e_i^2+\lambda_1|e_i|)}{\sum_i w_i}
        \right]\\
        &+\lambda_{\mathrm{TV}}\mathbb{E}\left[(1-s)
        \operatorname{TV}(\widetilde{P})\right].
    \end{split}
    \label{eq:flow_loss}
\end{equation}
Here, $w_i$ emphasizes active contact regions, and TV penalizes differences between adjacent rows and columns of $\widetilde{P}$. The factor $(1-s)$ upweights TV at low noise levels, where velocity errors perturb $\widetilde{P}$ less. We set $\lambda_1=0.5$, $\lambda_{\mathrm{TV}}=0.01$, and assign weight $\kappa=3$ to cells above $\tau_c=0.1$. At test time, we integrate eight noise samples from $s=1$ to $s=0$ via eight Euler steps of $v_\theta$ and average them to obtain $\widehat{P}_t$.

\subsection{Transferring the Prior to Robot Hands and Policies}
\begin{figure}[t]
    \centering
    \includegraphics[width=\columnwidth]{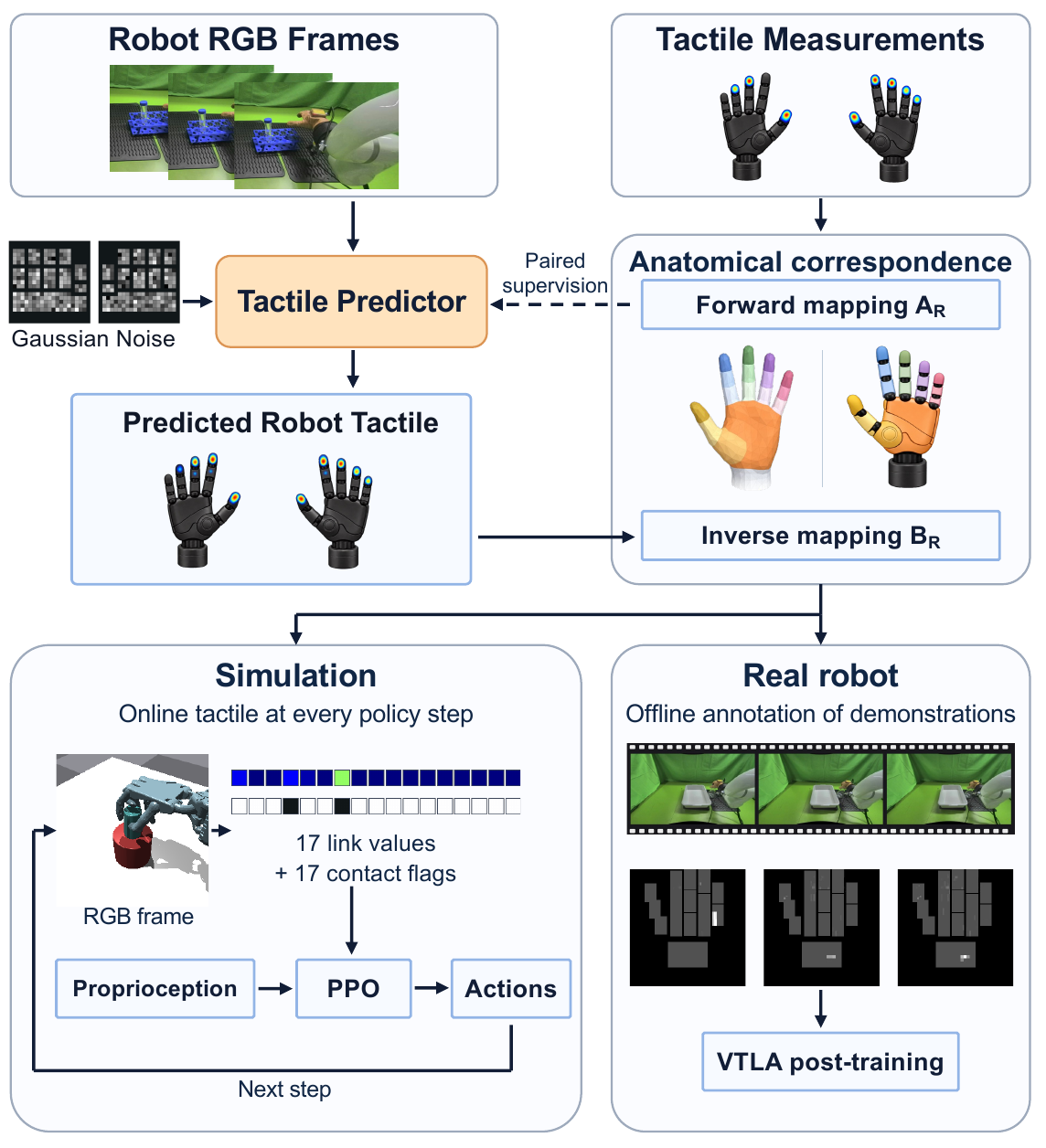}
    
    \vspace{-4mm}
    \caption{\textbf{Robot adaptation and policy learning.} Anatomical correspondences map robot measurements to the canonical grid for predictor fine-tuning under the same flow objective. The adapted predictor supplies touch information for robot policies in both the simulation and real robot hands.}
    \label{fig:robot_adaptation}
\end{figure}
\label{subsec:transfer}
\textbf{Shared tactile grid.}
All embodiments share a canonical $21\times21$ tactile grid for each hand (Fig.~\ref{fig:robot_adaptation}), organized into 16 anatomical regions that cover three segments per finger and the palm. These regions define correspondences with the sensing layout of each embodiment: every sensing unit of a robot hand is assigned to the region of the anatomical part it covers, and a unit without a region of its own, such as the little-finger metacarpal of the Shadow hand, shares the region of the part it belongs to. A validity mask excludes areas without sensor coverage from supervision. Keeping this grid as the output for every robot lets fine-tuning reuse the entire human-pretrained predictor with its output head, and only the mappings into and out of the grid depend on the embodiment. A forward mapping $A_R$ turns the measured signals $y_t^R$ of robot domain $R$ into normalized grid targets, and an inverse mapping $B_R$ turns the predicted grid into the tactile input of its policy:
\begin{equation}
    P_t^R=A_R(y_t^R),\qquad \widehat{z}_t^R=B_R\!\left(f_{\theta_R}(V_t^R)\right),
    \label{eq:robot_interface}
\end{equation}
where $V_t^R$ is the robot RGB clip and $\widehat{z}_t^R$ is the policy input.

\begin{figure}[t]
    \centering
    \includegraphics[width=\columnwidth]{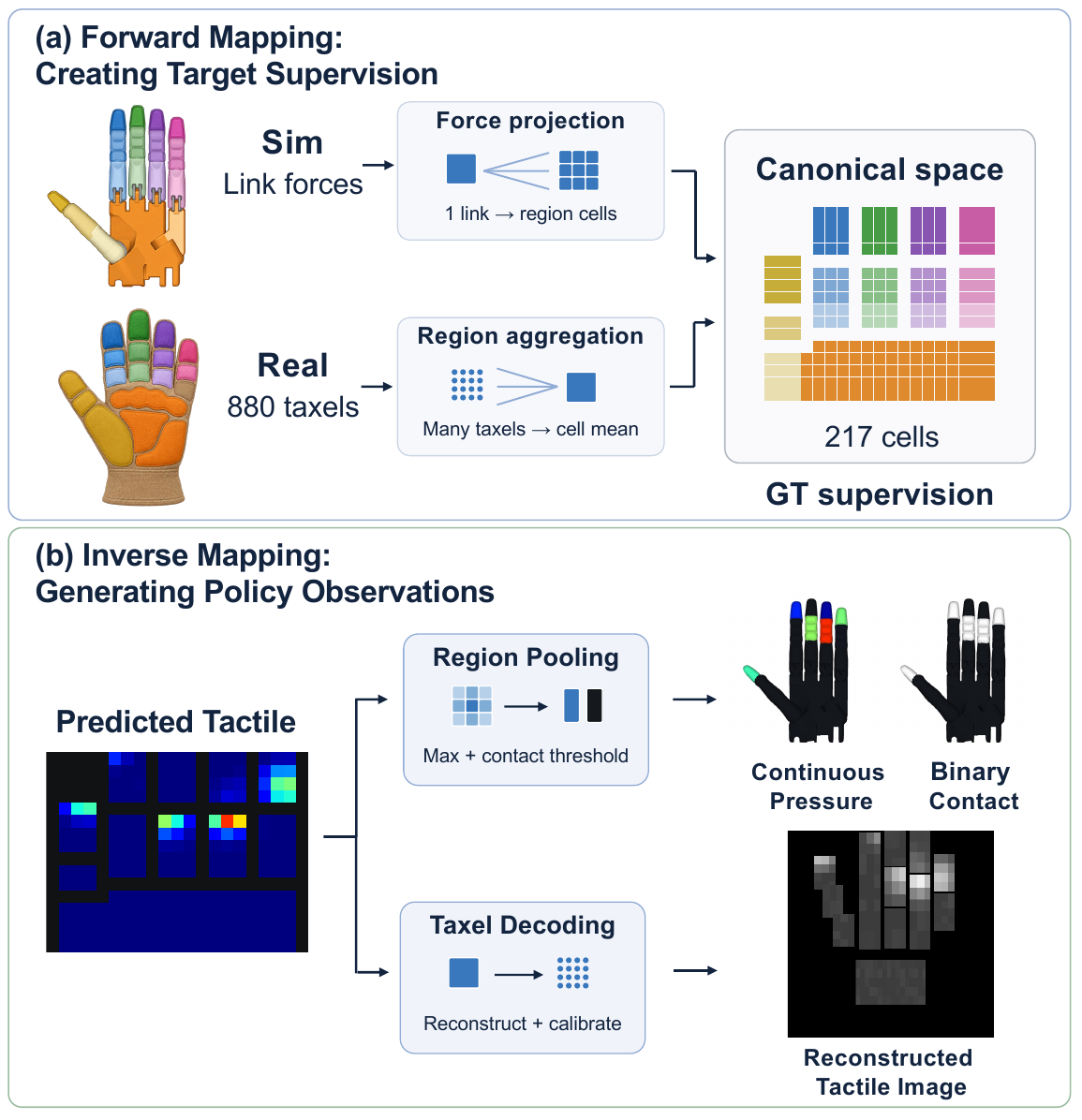}
    
    \vspace{-4mm}
    \caption{\textbf{Bidirectional anatomical mapping.} (a) The forward mapping creates fine-tuning supervision. Simulator link forces are distributed over their corresponding canonical cells, while real-glove taxels are averaged within each covered cell to form targets on the shared 217-cell grid. (b) The inverse mapping turns a predicted grid into policy inputs. Regional max pooling and thresholding produce a continuous pressure feature and a binary contact flag per link for reinforcement learning in simulation. A ridge decoder estimates the 880 taxel responses, which are quantile-mapped and rendered as a tactile image for $N_0$-VTLA post-training.}
    \label{fig:robot_anatomical_mapping}
\end{figure}

\textbf{Forward mapping and fine-tuning.}
In simulation (Fig.~\ref{fig:robot_anatomical_mapping}a), the Shadow hand provides 17 link-level contact forces. We spread each force over the cells of its anatomical region with Gaussian weights centered at the contact point on the link, with a standard deviation of one cell pitch. The weights are normalized to sum to one over the cells of that link, so each cell receives a share of the link force, and dividing by the cell sensing area gives pressure. Contact location thus remains visible on the grid, and the total force of every link is preserved. On the real robot, each hand wears a tactile glove with 15 sensing pads and 880 taxels, and the taxel readings assigned to each covered cell are averaged to form its target. In both domains, the targets are normalized by a per-mask maximum and clipped to $[0,1]$. Fine-tuning updates all trainable modules with the objective in~\eqref{eq:flow_loss} and supervises only the cells inside the validity mask, with both visual backbones frozen. Paired robot data come from successful simulator rollouts with recorded contact forces and from glove recordings of five real-robot tasks.

\textbf{Policy inputs for reinforcement learning.}
For PPO policies in simulation, $B_R$ runs online at every step and yields compact regional pressure and contact features (Fig.~\ref{fig:robot_anatomical_mapping}b). For each link $\ell$ with grid region $\mathcal{R}_\ell$, the maximum predicted pressure $q_{t,\ell}=\max_{i\in\mathcal{R}_\ell}\widehat{P}_{t,i}$ keeps localized contact peaks, and thresholding it gives a binary contact flag. The two channels receive separate weights:
\begin{equation}
    z_{t,\ell}^{\mathrm{pred}}=\bigl[\,w_c\,q_{t,\ell},\; w_b\,\mathbf{1}(q_{t,\ell}>\tau)\,\bigr].
    \label{eq:policy_tactile}
\end{equation}
We set the channel weights to $w_c=2$ and $w_b=3$ and the contact threshold to $\tau=0.1$. The 17 links give 17 continuous values and 17 binary flags per hand, which we append to the base observation of the policy.

\textbf{Policy inputs for VTLA post-training.}
On the real robot, we post-train $N_0$-VTLA~\cite{n0vtla}, which applies a frozen DINOv2 encoder to the difference between the current and first tactile images of an episode. To produce such images, we expand the 217-cell prediction back to the 880-taxel glove layout. A ridge-regression decoder fit on the collected paired data estimates the 880 taxel responses from the predicted cells and rescales them so that their average within each covered cell matches the predicted cell value. A task-specific quantile mapping then matches the decoded distribution to that of the measured glove, and the taxels are normalized and rendered onto a $224\times224$ grayscale canvas in the glove layout. Recorded glove data are rendered in the same format, so post-training with measured or predicted touch differs only in the source of the tactile images. The adapted predictor annotates every robot demonstration video, and the rendered images are stored with the original observations and action trajectories. During real-robot rollouts, the policy reads measured touch from the glove.

\section{Experiments}
\label{sec:experiments}
We test whether the predictor transfers to robot hands and whether its predicted touch improves their policies, first on human video (Section~\ref{subsec:exp_human}) and then on simulated and real robot hands (Sections~\ref{subsec:exp_sim}--\ref{subsec:exp_real}).

\begin{table*}[!b]
    \centering
    \caption{Human tactile prediction on EgoTouch. ``Ego2Contact (pose)'' conditions the generator on 3D hand pose tokens, whereas our full model conditions on hand-local visual features.}
    \label{tab:main_results}
    \vspace{-2mm}
    \small
    \begin{tabular*}{\textwidth}{@{\extracolsep{\fill}}lcccccccc@{}}
        \toprule
         & \multicolumn{4}{c}{Seen objects} & \multicolumn{4}{c}{Unseen objects}\\
        \cmidrule(lr){2-5}\cmidrule(lr){6-9}
        Method & C.IoU$\uparrow$ & V.IoU$\uparrow$ & T.Acc$\uparrow$ & MAE$\downarrow$ & C.IoU$\uparrow$ & V.IoU$\uparrow$ & T.Acc$\uparrow$ & MAE$\downarrow$\\
        \midrule
        TouchAnything~\cite{touchanything} & 0.5064 & 0.4260 & 0.8899 & 0.0485 & 0.4720 & \textbf{0.3457} & 0.8382 & 0.0641\\
        Ego2Contact (pose)                 & 0.5657 & 0.4899 & 0.9014 & 0.0385 & 0.4441 & 0.3183 & 0.8122 & 0.0596\\
        \textbf{Ego2Contact}               & \textbf{0.5735} & \textbf{0.4961} & \textbf{0.9018} & \textbf{0.0383} & \textbf{0.4812} & 0.3417 & \textbf{0.8544} & \textbf{0.0590}\\
        \bottomrule
    \end{tabular*}
\end{table*}

\subsection{Experimental Setup}
\label{subsec:exp_setup}
\textbf{Datasets.}
EgoTouch~\cite{touchanything} provides the human training data together with seen-object and unseen-object test sets of 179 and 85 episodes, respectively. EgoTactile~\cite{egotactile} serves as a second human domain, whose training split fine-tunes the predictor and whose object-held-out test set contains 55 episodes. EgoTactile records only the right hand, and several robot tasks use a single hand. In both cases, the tactile channels of the unused hand are set to zero.

\textbf{Metrics.}
Contact IoU (C.IoU) measures the overlap of binary contact maps, and volumetric IoU (V.IoU) measures the ratio of summed pressure intersection to union. Temporal accuracy (T.Acc) measures agreement in frame-level contact presence, and the mean absolute error (MAE) of normalized pressure is computed on valid sensor cells. For robot prediction, we also pool grid cells into the sensing links of the robot and report link-level contact IoU (Link C.IoU) and center-of-pressure error (Link CoP). For the correspondence ablation in Table~\ref{tab:correspondence_ablation}, grid-level metrics are not comparable across mappings because each mapping defines different grid targets, so we instead sum the predicted and ground-truth pressure within each link to obtain per-link net forces and compute C.IoU, V.IoU, and MAE on these 17 forces, normalized per task (Force C.IoU, Force V.IoU, Force MAE). These differ from Link C.IoU in table~\ref{tab:robot_posttrain}, which max-pools the cells of each link. Coarser mappings spread each force over more cells, which lowers the per-cell values that max-pooling reads but leaves the summed link force unchanged, so summation compares the mappings fairly. Robot policies are evaluated by task success rate.

\textbf{Simulated benchmarks.}
We study four contact-rich dexterous tasks in NVIDIA Isaac Gym~\cite{makoviychuk2021isaacgym}. Pen Cap Removal is built on Bi-DexHands~\cite{chen2022bidexhands}, and the other three tasks are adapted from VTDexManip~\cite{vtdexmanip}. All tasks use a simulated Shadow hand with full-state proprioceptive observations, and each policy is trained with PPO~\cite{schulman2017ppo} for 2,500 to 12,000 iterations depending on the task. We run 200 parallel environments for Base and Base+GT conditions, and 60 for the predicted-touch condition since it queries the predictor online at every step, with its iteration count scaled so that the total number of environment steps matches. The three conditions differ only in the tactile channel of the PPO observation. Base uses the observations defined by each benchmark, and Base+GT and Base+Pred append 17 pressure values and 17 binary contact flags per hand computed from simulator contact forces and from the adapted predictor, respectively.

\subsection{Tactile Prediction on Human Video}
\label{subsec:exp_human}
Ego2Contact predicts human touch more accurately than both baselines on seven of the eight metrics (Table~\ref{tab:main_results}). Compared with our reproduction of the regression predictor TouchAnything~\cite{touchanything}, it achieves the highest contact IoU and temporal accuracy and the lowest pressure MAE on both object splits, with larger gains on seen objects. The full model therefore localizes contact and tracks its timing better than prior work on both object splits.

Hand-local appearance generalizes to unseen objects better than 3D hand pose. Conditioning the generator on pose tokens raises seen-object contact IoU over TouchAnything from 0.5064 to 0.5657 but lowers unseen-object contact IoU from 0.4720 to 0.4441. Replacing pose with hand-local visual features recovers this loss and improves all four unseen-object metrics. Pose thus fits the hand configurations seen in training, whereas hand-local appearance carries over to new objects and drives the gains of the full model.

\begin{figure*}[t]
    \centering
    \includegraphics[width=0.92\textwidth]{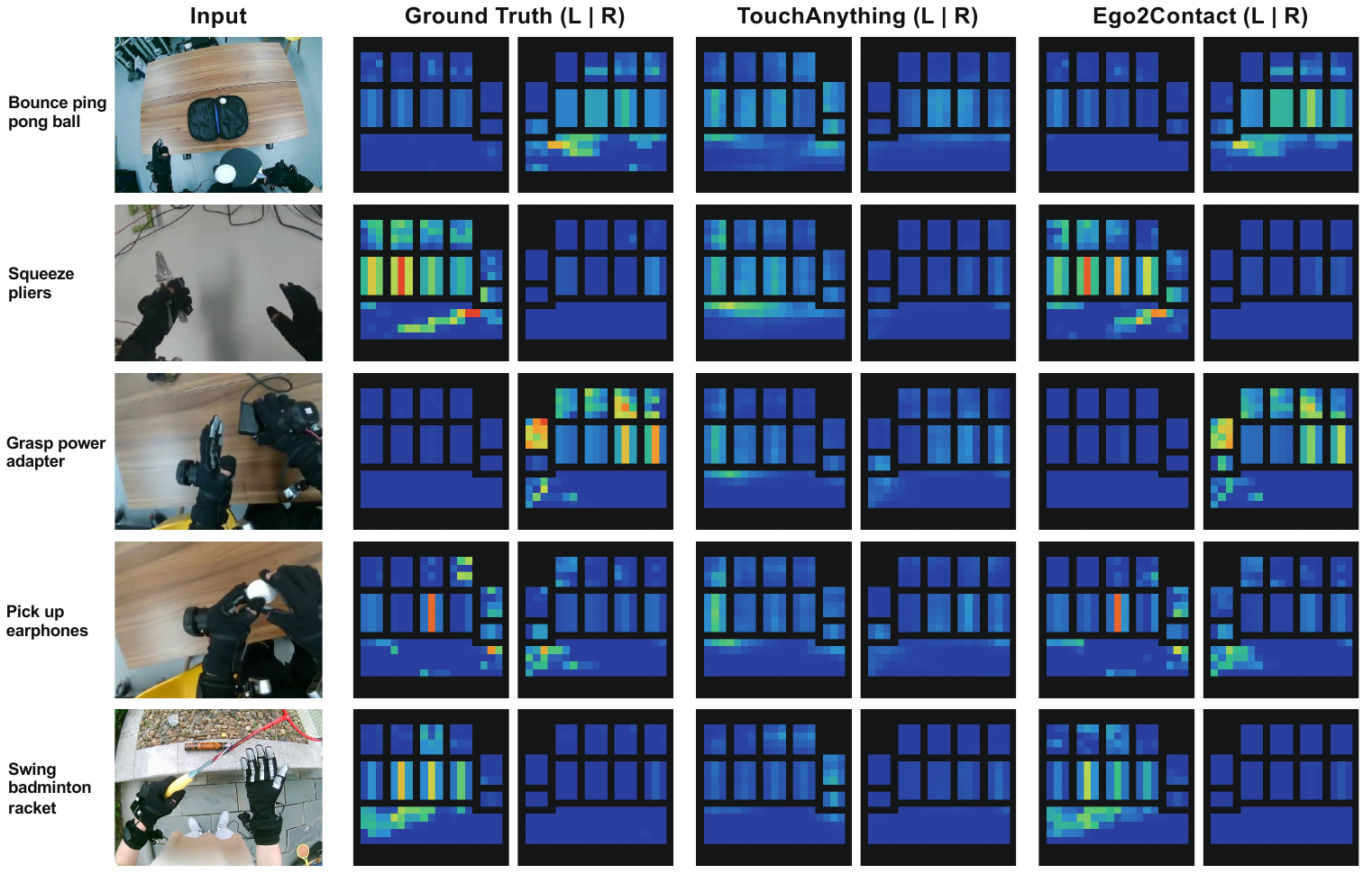}
    
    \vspace{-4mm}
    \caption{\textbf{Qualitative human tactile prediction.} For representative EgoTouch interactions (rows), we show the egocentric input and the left and right tactile maps of Ego2Contact and TouchAnything next to the ground truth. Ego2Contact localizes contact and relative intensity more faithfully.}
    \label{fig:human_tactile_viz}
\end{figure*}

The advantage of hand-local appearance also holds on a second tactile dataset. After fine-tuning both variants on EgoTactile from their human-trained checkpoints, Ego2Contact outperforms the pose variant on all four metrics on the object-held-out split, raising contact IoU from 0.6414 to 0.6988 (Table~\ref{tab:egotactile}). Fig.~\ref{fig:human_tactile_viz} shows the same trend qualitatively.

\begin{table}[tb]
    \centering
    \caption{Cross-dataset adaptation on EgoTactile (object-held-out split, right hand only).}
    \label{tab:egotactile}
    \vspace{-2mm}
    \small
    \begin{tabular*}{\columnwidth}{@{\extracolsep{\fill}}lcccc@{}}
        \toprule
        Method & C.IoU$\uparrow$ & V.IoU$\uparrow$ & T.Acc$\uparrow$ & MAE$\downarrow$\\
        \midrule
        Ego2Contact (pose) & 0.6414 & 0.4557 & 0.8520 & 0.0352\\
        \textbf{Ego2Contact} & \textbf{0.6988} & \textbf{0.5453} & \textbf{0.9209} & \textbf{0.0304}\\
        \bottomrule
    \end{tabular*}
\end{table}

\subsection{Transfer to Simulated Robot Hands}
\label{subsec:exp_sim}
To adapt the predictor to the simulated hand, we collect paired RGB videos and link-level contact forces from successful rollouts of ground-truth-touch policies. To account for varying episode lengths, we sample a task-balanced set of 70k frames to fine-tune TouchAnything and Ego2Contact from their human-trained checkpoints, evaluating on 249 held-out episodes. After adaptation, Ego2Contact localizes robot contact more accurately than TouchAnything (Table~\ref{tab:robot_posttrain}). It raises contact IoU from 0.4154 to 0.4756 on the grid and from 0.5200 to 0.5709 on the robot links, and lowers the center-of-pressure error from 0.2999 to 0.2242, while pressure MAE and temporal accuracy stay comparable. The accuracy gain on human data therefore carries over to robot hands, mainly in where contact occurs.

\begin{table}[t]
    \centering
    \caption{Robot tactile prediction after adaptation to the simulated Shadow hand, averaged over all held-out episodes. Link metrics max-pool the grid cells of each link.}
    \label{tab:robot_posttrain}
    \vspace{-2mm}
    \footnotesize
    \setlength{\tabcolsep}{2pt}
    \begin{tabular*}{\columnwidth}{@{\extracolsep{\fill}}lcccccc@{}}
        \toprule
         & \multicolumn{4}{c}{Grid} & \multicolumn{2}{c}{Link}\\
        \cmidrule(lr){2-5}\cmidrule(lr){6-7}
        Method & C.IoU$\uparrow$ & V.IoU$\uparrow$ & MAE$\downarrow$ & T.Acc$\uparrow$ & C.IoU$\uparrow$ & CoP$\downarrow$\\
        \midrule
        TouchAnything~\cite{touchanything} & 0.4154 & 0.3843 & \textbf{0.0025} & \textbf{0.9923} & 0.5200 & 0.2999\\
        \textbf{Ego2Contact} & \textbf{0.4756} & \textbf{0.4403} & \textbf{0.0025} & 0.9907 & \textbf{0.5709} & \textbf{0.2242}\\
        \bottomrule
    \end{tabular*}
\end{table}

Table~\ref{tab:correspondence_ablation} studies the anatomical correspondence. Coarser correspondences cannot represent link-level contact, since the forces of merged links become indistinguishable on the grid: one region per finger (6 regions) lowers Force C.IoU from 0.5554 to 0.2037, and merging the middle and proximal segments (12 regions) lowers it to 0.4504. At 16 regions, spreading each force uniformly over its region instead of around the contact point lowers Force C.IoU by 0.02 and Force V.IoU by 0.03. Replacing the grid with a direct 17-link output head on the same pretrained transformer gives 0.5504, on par with our grid mapping. The contact-centered Gaussian mapping therefore lets the shared grid match the accuracy of a native per-link output while keeping one representation for human, simulated, and real hands.

\begin{table}[t]
    \centering
    \caption{Ablation of the anatomical correspondence on the simulated Shadow hand over held-out episodes. Metrics use per-link forces, the sum of the grid cells of each link (Sec. IV-A). All rows fine-tune the same pretrained predictor.}
    \label{tab:correspondence_ablation}
    \vspace{-2mm}
    \small
    \setlength{\tabcolsep}{6pt}
    \begin{tabular*}{\columnwidth}{@{\extracolsep{\fill}}lccc@{}}
        \toprule
         & \multicolumn{3}{c}{Per-link force}\\
        \cmidrule(lr){2-4}
        Correspondence & C.IoU$\uparrow$ & V.IoU$\uparrow$ & MAE$\downarrow$\\
        \midrule
        6 regions (per finger) & 0.2037 & 0.1911 & 0.0208\\
        12 regions (tip / rest) & 0.4504 & 0.3905 & 0.0138\\
        16 regions, uniform & 0.5355 & 0.4583 & 0.0117\\
        \textbf{16 regions, Gaussian (ours)} & \textbf{0.5554} & \textbf{0.4842} & \textbf{0.0114}\\
        \midrule
        Direct 17-link head (no grid) & 0.5504 & 0.4777 & 0.0117\\
        \bottomrule
    \end{tabular*}
\end{table}

Predicted touch improves policy success in simulation on three of the four tasks (Table~\ref{tab:policy_results}), raising average success from 44.4\% without touch to 48.5\%. It approaches ground-truth touch on Bottle Cap Turning (71.0\% versus 74.1\%) and recovers a smaller share of its gain on Lever Sliding and Object Handover. On Pen Cap Removal, ground-truth touch reaches 95.0\%, whereas predicted touch stays near the no-touch baseline because contact points are fully occluded and the active finger is visually ambiguous. Predicted touch thus helps when the contact region is visible to the camera, and heavy occlusion remains the main failure case.

\begin{table}[t]
    \centering
    \caption{Policy success rate (\%) in simulation, averaged over four seeds. Base uses no tactile input, and Base+Pred and Base+GT append predicted and ground-truth touch online.}
    \label{tab:policy_results}
    \vspace{-2mm}
    \small
    \begin{tabular*}{\columnwidth}{@{\extracolsep{\fill}}lccc@{}}
        \toprule
        Task & Base & Base+Pred & Base+GT\\
        \midrule
        Pen cap removal & 0.0 & 1.0 & 95.0\\
        Lever sliding & 84.0 & 88.8 & 98.2\\
        Bottle cap turning & 64.1 & 71.0 & 74.1\\
        Object handover & 29.3 & 33.1 & 49.1\\
        \bottomrule
    \end{tabular*}
\end{table}

\subsection{Transfer to Real Robot Hands}
\label{subsec:exp_real}
To test real hardware transfer, we design four single-arm and bimanual manipulation tasks on a dual-arm xArm6 system with BrainCo Revo 2 hands. Each hand wears a tactile glove with 880 taxels (15 sensing pads over the fingers and palm), observed by a head RGB-D camera and two wrist cameras. Expert demonstrations are collected via VR teleoperation. We adapt the human-trained predictor using roughly 123k paired visual--tactile frames from five tasks disjoint from the post-training demonstrations. On held-out recordings, it reaches 0.27 contact IoU, 0.19 volumetric IoU, and 0.97 temporal accuracy. The adapted predictor then annotates demonstration videos for $N_0$-VTLA post-training. Sharing all non-tactile inputs, the Pred, Zero, and GT conditions use predicted tactile images, zeros, and glove measurements, respectively. During rollouts, every policy receives glove measurements, so the comparison isolates the effect of the tactile annotations used in post-training. Rollouts run at about 24.5 Hz with a limit of 60 seconds for single-arm tasks and 90 seconds for the bimanual task.

\begin{figure}[t]
    \centering
    \includegraphics[width=\columnwidth]{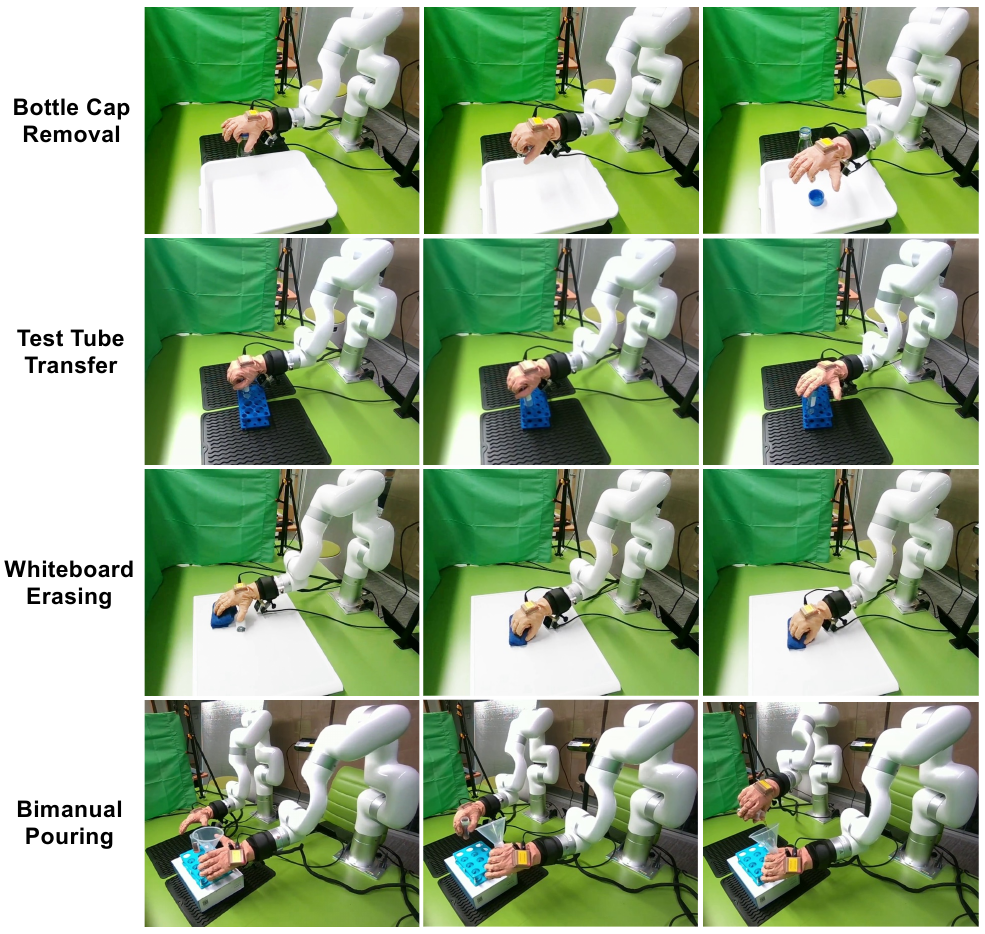}
    
    \vspace{-4mm}
    \caption{\textbf{Real-world task execution.} Representative frames of the four real-robot tasks, with each row showing one task progressing from left to right.}
    \label{fig:real_world_deployment}
\end{figure}

Post-training with predicted touch improves real-robot success on all four tasks (Table~\ref{tab:real_results}). Averaged over the tasks, success rises from 27.5\% with zero-valued touch to 45.0\% with predicted touch, while measured touch reaches 55.0\%. Predicted touch matches measured touch on Bottle Cap Removal and comes within one success on Test Tube Transfer and Whiteboard Erasing, while the largest gap remains on Bimanual Pouring. Fig.~\ref{fig:real_world_deployment} shows execution sequences for all four tasks. Touch predicted from video therefore recovers nearly two thirds of the gain from measured touch when demonstrations lack tactile recordings.

\begin{table}[t]
    \centering
    \caption{Real-robot successes over 10 trials per task. Pred uses predicted tactile images in post-training, and Zero and GT replace them with zero values and glove measurements. All rollouts use measured touch.}
    \label{tab:real_results}
    \vspace{-2mm}
    \small
    \begin{tabular*}{\columnwidth}{@{\extracolsep{\fill}}lccc@{}}
        \toprule
        Task & Zero & Pred & GT\\
        \midrule
        Bottle cap removal & 5/10 & 7/10 & 7/10\\
        Test tube transfer & 3/10 & 5/10 & 6/10\\
        Whiteboard erasing & 2/10 & 3/10 & 4/10\\
        Bimanual pouring & 1/10 & 3/10 & 5/10\\
        \bottomrule
    \end{tabular*}
\end{table}

\section{Conclusion}
We presented Ego2Contact, which learns a visual--tactile prior from egocentric human interactions and transfers it to robot hands. Temporal scene context and hand-local visual features condition a flow-matching tactile generator, and anatomical region correspondences connect its canonical grid to different robot sensing layouts. Ego2Contact outperforms regression and pose-conditioned baselines on human data and improves contact localization on simulated robot hands after adaptation. Used as online observations for simulated RL and offline annotations for real-robot VTLA post-training, predicted touch consistently improves manipulation success over no-touch baselines in both simulation and real-world deployments. One limitation is that the fixed grid discards sub-cell detail for dense sensors. Future work will learn a shared tactile representation that preserves the native sensor resolution and extends to other sensor types such as vision-based tactile sensors.

\bibliographystyle{IEEEtran}
\bibliography{bibliography.bib}

@IEEEtranBSTCTL{IEEEexample:BSTcontrol,
  CTLuse_url = "no"
}

@INPROCEEDINGS{pressurevision,
  author={Grady, Patrick and Tang, Chengcheng and Brahmbhatt, Samarth and Twigg, Christopher D. and Wan, Chengde and Hays, James and Kemp, Charles C.},
  booktitle={Computer Vision -- ECCV 2022},
  title={{PressureVision}: Estimating Hand Pressure from a Single {RGB} Image},
  year={2022},
  pages={328--345},
  doi={10.1007/978-3-031-20068-7_19}
}

@INPROCEEDINGS{pressurevisionpp,
  author={Grady, Patrick and Collins, Jeremy A. and Tang, Chengcheng and Twigg, Christopher D. and Aneja, Kunal and Hays, James and Kemp, Charles C.},
  booktitle={Proceedings of the IEEE/CVF Winter Conference on Applications of Computer Vision (WACV)},
  title={{PressureVision++}: Estimating Fingertip Pressure from Diverse {RGB} Images},
  year={2024},
  pages={8698--8708}
}

@INPROCEEDINGS{haco,
  author={Jung, Daniel Sungho and Lee, Kyoung Mu},
  booktitle={Advances in Neural Information Processing Systems},
  title={Learning Dense Hand Contact Estimation from Imbalanced Data},
  year={2025},
  volume={38},
  doi={10.52202/085713-4017}
}

@MISC{hope,
  author={Jeon, Subin and Kim, Byungjun and Joo, Hanbyul},
  title={{HOPE}: Hand-Object Pressure Estimation from Monocular Videos},
  year={2026},
  eprint={2608.06192},
  archivePrefix={arXiv},
  primaryClass={cs.CV},
  howpublished={\url{https://arxiv.org/abs/2608.06192}}
}

@INPROCEEDINGS{egophi,
  author={Ilic, Andela and Schuchert, Rachel and Jiang, Yijing and Holz, Christian},
  booktitle={Computer Vision -- ECCV 2026},
  title={{EgoPHI}: Estimating 3{D} Hand-Object Contact and Force from Egocentric Vision},
  year={2026},
  pages={410--428}
}

@article{johansson2009tactile,
  author = {Johansson, Roland S. and Flanagan, J. Randall},
  title = {Coding and use of tactile signals from the fingertips in object manipulation tasks},
  journal = {Nature Reviews Neuroscience},
  volume = {10},
  number = {5},
  pages = {345--359},
  year = {2009},
  doi = {10.1038/nrn2621},
  url = {https://www.nature.com/articles/nrn2621}
}

@article{calandra2018feeling,
  author = {Calandra, Roberto and Owens, Andrew and Jayaraman, Dinesh and Lin, Justin and Yuan, Wenzhen and Malik, Jitendra and Adelson, Edward H. and Levine, Sergey},
  title = {More Than a Feeling: Learning to Grasp and Regrasp using Vision and Touch},
  journal = {IEEE Robotics and Automation Letters},
  volume = {3},
  number = {4},
  pages = {3300--3307},
  year = {2018},
  url = {https://arxiv.org/abs/1805.11085}
}

@article{dahiya2010tactile,
  author = {Dahiya, Ravinder S. and Metta, Giorgio and Valle, Maurizio and Sandini, Giulio},
  title = {Tactile Sensing---From Humans to Humanoids},
  journal = {IEEE Transactions on Robotics},
  volume = {26},
  number = {1},
  pages = {1--20},
  year = {2010},
  doi = {10.1109/TRO.2009.2033627},
  url = {https://ieeexplore.ieee.org/document/5339133}
}

@inproceedings{egopressure,
  author = {Zhao, Yiming and Kwon, Taein and Streli, Paul and Pollefeys, Marc and Holz, Christian},
  title = {{EgoPressure}: A Dataset for Hand Pressure and Pose Estimation in Egocentric Vision},
  booktitle = {Proceedings of the IEEE/CVF Conference on Computer Vision and Pattern Recognition (CVPR)},
  year = {2025},
  pages = {27727--27738},
  doi = {10.1109/CVPR52734.2025.02582},
  url = {https://siplab.org/projects/EgoPressure}
}

@misc{egopressdiff,
  author = {Zeng, Yuan and Gao, Zilue and Shi, Yujia and Lu, Zongqing and Yang, Wenming and Liao, QingMin},
  title = {{EgoPressDiff}: Multimodal Video Diffusion for Egocentric {UV}-Domain Hand-Pressure Estimation},
  year = {2026},
  eprint = {2606.06872},
  archivePrefix = {arXiv},
  url = {https://arxiv.org/abs/2606.06872}
}

@misc{touchanything,
  author = {Zhou, Jianyi and Gao, Ziteng and Hong, Feiyang and Liu, Zirui and Zhang, Guannan and Dai, Weisheng and Zhen, Ruichen and Lyu, Chuqiao and Wu, Haotian and Mao, Yinian and Wang, Xushi and Jiang, Yuxiang and Ding, Wenbo and Yang, Shuo},
  title = {{TouchAnything}: A Dataset and Framework for Bimanual Tactile Estimation from Egocentric Video},
  year = {2026},
  eprint = {2605.13083},
  archivePrefix = {arXiv},
  url = {https://arxiv.org/abs/2605.13083}
}

@inproceedings{egotactile,
  author = {Zeng, Yuan and Shi, Yujia and Tan, Tiao and Li, Xingting and Qin, Yaqi and Lu, Zongqing and Yang, Wenming and Xue, Jing-Hao and Liao, Qingmin},
  title = {{EgoTactile}: Learning Grasp Pressure for Everyday Objects from Egocentric Video},
  booktitle = {International Conference on Machine Learning (ICML)},
  year = {2026},
  url = {https://arxiv.org/abs/2606.09243}
}

@misc{egotac,
  author = {Zhang, Wenkang and Yuan, Chengbo and Zhang, Zicheng and Cheng, Zhengxue and Gao, Yang},
  title = {{EgoTac}: In-the-wild Tactile Prediction from Egocentric Vision},
  year = {2026},
  eprint = {2608.15060},
  archivePrefix = {arXiv},
  url = {https://arxiv.org/abs/2608.15060}
}

@misc{htbench,
  author = {Huang, Yuzhe and Wu, Jiaping and Jiang, Jiaming and Lin, Hezhe and Aierken, Aikebaier and Wang, Yunlong and Cheng, Kun and Li, Wanlin and Xiao, Chenxi and Jiao, Ziyuan and Zhong, Yuanxin},
  title = {{HT-Bench}: Benchmarking and Learning Dexterous Full-Hand Tactile Representations with Egocentric Vision},
  year = {2026},
  eprint = {2606.19161},
  archivePrefix = {arXiv},
  url = {https://arxiv.org/abs/2606.19161v2}
}

@misc{opentouch,
  author = {Song, Yuxin Ray and Li, Jinzhou and Fu, Rao and Murphy, Devin and Zhou, Kaichen and Shiv, Rishi and Li, Yaqi and Xiong, Haoyu and Owens, Crystal Elaine and Du, Yilun and Luo, Yiyue and Cheng, Xianyi and Torralba, Antonio and Matusik, Wojciech and Liang, Paul Pu},
  title = {{OPENTOUCH}: Bringing Full-Hand Touch to Real-World Interaction},
  year = {2025},
  eprint = {2512.16842},
  archivePrefix = {arXiv},
  url = {https://arxiv.org/abs/2512.16842}
}

@misc{tacimag,
  author = {Zhang, Zhiyuan and Desai, Adeesh and Hu, Jyun-Chi and Saka, Yosuke and Luu, Quan Khanh and Lei, Jiuzhou and Soleymanzadeh, Davood and Zhang, Bihao and Zheng, Minghui and She, Yu},
  title = {Imagining the Sense of Touch: Touch-Informed Manipulation via Imagined Tactile Representations},
  year = {2026},
  eprint = {2607.01684},
  archivePrefix = {arXiv},
  url = {https://arxiv.org/abs/2607.01684}
}

@misc{htac,
  author = {Zhang, Chi and Cai, Penglin and Xi, Ziheng and Yuan, Haoqi and Luo, Hao and Zhang, Wanpeng and Zheng, Sipeng and Xu, Chaoyi and Lu, Zongqing},
  title = {Human-Centric Transferable Tactile Pre-Training for Dexterous Robotic Manipulation},
  year = {2026},
  eprint = {2607.01067},
  archivePrefix = {arXiv},
  url = {https://arxiv.org/abs/2607.01067}
}

@misc{touchworld,
  author = {Zhou, Jianyi and Hong, Feiyang and Li, Yunhao and Zhao, Yicheng and Cen, Yongjue and Liu, Zirui and Huang, Jiakang and Chen, Zirui and Zhang, Ruiyang and Zhu, Weizhuo and Song, Xuhua and Yang, Shuo},
  title = {{TouchWorld}: A Predictive and Reactive Tactile Foundation Model for Dexterous Manipulation},
  year = {2026},
  eprint = {2607.07287},
  archivePrefix = {arXiv},
  url = {https://arxiv.org/abs/2607.07287}
}

@misc{vitacworld,
  author = {Huang, Yunao and Sang, Shiyu and Ni, Suting and Lu, Haotao and Wu, Shijie and Lv, Rui and Guo, Ziyang and Shi, Ye and Wang, Jingya},
  title = {{ViTacWorld}: Scaling Visuo-Tactile World Models for Contact-Rich Robot Manipulation},
  year = {2026},
  eprint = {2607.22530},
  archivePrefix = {arXiv},
  url = {https://arxiv.org/abs/2607.22530}
}

@inproceedings{brahmbhatt2019contactdb,
  author = {Brahmbhatt, Samarth and Ham, Cusuh and Kemp, Charles C. and Hays, James},
  title = {{ContactDB}: Analyzing and Predicting Grasp Contact via Thermal Imaging},
  booktitle = {Proceedings of the IEEE/CVF Conference on Computer Vision and Pattern Recognition (CVPR)},
  year = {2019},
  pages = {8709--8719},
  url = {https://arxiv.org/abs/1904.06830}
}

@inproceedings{taheri2020grab,
  author = {Taheri, Omid and Ghorbani, Nima and Black, Michael J. and Tzionas, Dimitrios},
  title = {{GRAB}: A Dataset of Whole-Body Human Grasping of Objects},
  booktitle = {European Conference on Computer Vision (ECCV)},
  year = {2020},
  url = {https://grab.is.tue.mpg.de/}
}

@inproceedings{reactivediffusionpolicy,
  author = {Xue, Han and Ren, Jieji and Chen, Wendi and Zhang, Gu and Fang, Yuan and Gu, Guoying and Xu, Huazhe and Lu, Cewu},
  title = {{Reactive Diffusion Policy}: Slow-Fast Visual-Tactile Policy Learning for Contact-Rich Manipulation},
  booktitle = {Proceedings of Robotics: Science and Systems (RSS)},
  year = {2025},
  doi = {10.15607/RSS.2025.XXI.052},
  url = {https://www.roboticsproceedings.org/rss21/p052.html}
}

@misc{rlhnd,
  author = {Moon, Seungjun and Jeon, Subin and Kim, Sangwoo and Joo, Hanbyul and Shin, Jinwoo},
  title = {{RLHND}: Video Foundation Models as Physically Grounded Hand Trackers for Robot Learning},
  year = {2026},
  eprint = {2610.09455},
  archivePrefix = {arXiv},
  howpublished = {\url{https://arxiv.org/abs/2610.09455}}
}

@inproceedings{vtdexmanip,
  author = {Liu, Qingtao and Cui, Yu and Sun, Zhengnan and Li, Gaofeng and Chen, Jiming and Ye, Qi},
  title = {{VTDexManip}: A Dataset and Benchmark for Visual-Tactile Pretraining and Dexterous Manipulation with Reinforcement Learning},
  booktitle = {International Conference on Learning Representations (ICLR)},
  year = {2025}
}

@article{gelsight,
  author = {Yuan, Wenzhen and Dong, Siyuan and Adelson, Edward H.},
  title = {{GelSight}: High-Resolution Robot Tactile Sensors for Estimating Geometry and Force},
  journal = {Sensors},
  volume = {17},
  number = {12},
  pages = {2762},
  year = {2017},
  doi = {10.3390/s17122762}
}

@article{dinov2,
  author = {Oquab, Maxime and Darcet, Timothée and Moutakanni, Théo and Vo, Huy and Szafraniec, Marc and Khalidov, Vasil and others},
  title = {{DINOv2}: Learning Robust Visual Features without Supervision},
  journal = {Transactions on Machine Learning Research (TMLR)},
  year = {2024}
}

@inproceedings{wilor,
  author = {Potamias, Rolandos Alexandros and Zhang, Jinglei and Deng, Jiankang and Zafeiriou, Stefanos},
  title = {{WiLoR}: End-to-end 3{D} Hand Localization and Reconstruction in-the-wild},
  booktitle = {Proceedings of the IEEE/CVF Conference on Computer Vision and Pattern Recognition (CVPR)},
  year = {2025}
}

@inproceedings{dit,
  author = {Peebles, William and Xie, Saining},
  title = {Scalable Diffusion Models with Transformers},
  booktitle = {Proceedings of the IEEE/CVF International Conference on Computer Vision (ICCV)},
  year = {2023}
}

@inproceedings{rectifiedflow,
  author = {Liu, Xingchao and Gong, Chengyue and Liu, Qiang},
  title = {Flow Straight and Fast: Learning to Generate and Transfer Data with Rectified Flow},
  booktitle = {International Conference on Learning Representations (ICLR)},
  year = {2023}
}

@inproceedings{lee2019makingsense,
  author = {Lee, Michelle A. and Zhu, Yuke and Srinivasan, Krishnan and Shah, Parth and Savarese, Silvio and Fei-Fei, Li and Garg, Animesh and Bohg, Jeannette},
  title = {Making Sense of Vision and Touch: Self-Supervised Learning of Multimodal Representations for Contact-Rich Tasks},
  booktitle = {IEEE International Conference on Robotics and Automation (ICRA)},
  year = {2019}
}

@inproceedings{guzey2023dexterity,
  author = {Güzey, Irmak and Evans, Ben and Chintala, Soumith and Pinto, Lerrel},
  title = {Dexterity from Touch: Self-Supervised Pre-Training of Tactile Representations with Robotic Play},
  booktitle = {Conference on Robot Learning (CoRL)},
  year = {2023}
}

@inproceedings{nair2022r3m,
  author = {Nair, Suraj and Rajeswaran, Aravind and Kumar, Vikash and Finn, Chelsea and Gupta, Abhinav},
  title = {{R3M}: A Universal Visual Representation for Robot Manipulation},
  booktitle = {Conference on Robot Learning (CoRL)},
  year = {2022}
}

@inproceedings{qin2022dexmv,
  author = {Qin, Yuzhe and Wu, Yueh-Hua and Liu, Shaowei and Jiang, Hanwen and Yang, Ruihan and Fu, Yang and Wang, Xiaolong},
  title = {{DexMV}: Imitation Learning for Dexterous Manipulation from Human Videos},
  booktitle = {European Conference on Computer Vision (ECCV)},
  year = {2022}
}

@inproceedings{qi2023rotateit,
  author = {Qi, Haozhi and Yi, Brent and Suresh, Sudharshan and Lambeta, Mike and Ma, Yi and Calandra, Roberto and Malik, Jitendra},
  title = {General In-Hand Object Rotation with Vision and Touch},
  booktitle = {Conference on Robot Learning (CoRL)},
  year = {2023}
}

@inproceedings{shaw2022videodex,
  author = {Shaw, Kenneth and Bahl, Shikhar and Pathak, Deepak},
  title = {{VideoDex}: Learning Dexterity from Internet Videos},
  booktitle = {Conference on Robot Learning (CoRL)},
  year = {2022}
}

@inproceedings{wang2024dexcap,
  author = {Wang, Chen and Shi, Haochen and Wang, Weizhuo and Zhang, Ruohan and Fei-Fei, Li and Liu, C. Karen},
  title = {{DexCap}: Scalable and Portable Mocap Data Collection System for Dexterous Manipulation},
  booktitle = {Robotics: Science and Systems (RSS)},
  year = {2024}
}

@inproceedings{chen2022bidexhands,
  author = {Chen, Yuanpei and Wu, Tianhao and Wang, Shengjie and Feng, Xidong and Jiang, Jiechuang and McAleer, Stephen Marcus and Geng, Yiran and Dong, Hao and Lu, Zongqing and Zhu, Song-Chun and Yang, Yaodong},
  title = {Towards Human-Level Bimanual Dexterous Manipulation with Reinforcement Learning},
  booktitle = {Advances in Neural Information Processing Systems (NeurIPS) Datasets and Benchmarks Track},
  year = {2022}
}

@misc{schulman2017ppo,
  author = {Schulman, John and Wolski, Filip and Dhariwal, Prafulla and Radford, Alec and Klimov, Oleg},
  title = {Proximal Policy Optimization Algorithms},
  year = {2017},
  eprint = {1707.06347},
  archivePrefix = {arXiv},
  howpublished = {\url{https://arxiv.org/abs/1707.06347}}
}

@inproceedings{makoviychuk2021isaacgym,
  author = {Makoviychuk, Viktor and Wawrzyniak, Lukasz and Guo, Yunrong and Lu, Michelle and Storey, Kier and Macklin, Miles and Hoeller, David and Rudin, Nikita and Allshire, Arthur and Handa, Ankur and State, Gavriel},
  title = {{Isaac Gym}: High Performance {GPU}-Based Physics Simulation for Robot Learning},
  booktitle = {Advances in Neural Information Processing Systems (NeurIPS) Datasets and Benchmarks Track},
  year = {2021}
}

@misc{n0vtla,
  author = {{NeoteAI Team} and {Fudan TEAI Team}},
  title = {{$N_0$-VTLA}: Scaling Vision-Tactile-Language-Action Model with Latent Tactile Tokens},
  year = {2026},
  eprint = {2607.23782},
  archivePrefix = {arXiv},
  howpublished = {\url{https://arxiv.org/abs/2607.23782}}
}

@inproceedings{epickitchens,
  author    = {Damen, D. and Doughty, H. and Farinella, G. M. and Fidler, S. and Furnari, A. and Kazakos, E. and Moltisanti, D. and Munro, J. and Perrett, T. and Price, W. and Wray, M.},
  title     = {Scaling egocentric vision: The {EPIC-KITCHENS} dataset},
  booktitle = {Proceedings of the European Conference on Computer Vision (ECCV)},
  year      = {2018},
  url       = {https://openaccess.thecvf.com/content_ECCV_2018/html/Dima_Damen_Scaling_Egocentric_Vision_ECCV_2018_paper.html}
}

@inproceedings{ego4d,
  author    = {Grauman, K. and others},
  title     = {{Ego4D}: Around the world in 3,000 hours of egocentric video},
  booktitle = {Proceedings of the IEEE/CVF Conference on Computer Vision and Pattern Recognition (CVPR)},
  year      = {2022}
}

@inproceedings{egodex,
  author    = {Hoque, R. and Huang, P. and Yoon, D. J. and Sivapurapu, M. and Zhang, J.},
  title     = {{EgoDex}: Learning dexterous manipulation from large-scale egocentric video},
  booktitle = {International Conference on Learning Representations (ICLR)},
  year      = {2026},
  url       = {https://arxiv.org/abs/2505.11709}
}

\end{document}